\documentclass{article}

\usepackage{microtype}
\usepackage{graphicx}
\usepackage{subfigure}
\usepackage{booktabs} % for professional tables
\usepackage{natbib}
\usepackage{hyperref}

\usepackage[accepted]{icml2023}

\usepackage{amsmath}
\usepackage{amssymb}
\usepackage{mathtools}
\usepackage{amsthm}

\usepackage[capitalize,noabbrev]{cleveref}

\theoremstyle{plain}

\theoremstyle{definition}

\theoremstyle{remark}

\usepackage[textsize=tiny]{todonotes}

\icmltitlerunning{Neural Loss Function Evolution}

\begin{document}

\twocolumn[
\icmltitle{Neural Loss Function Evolution for Large-Scale \\ 
Image Classifier Convolutional Neural Networks}

% It is OKAY to include author information, even for blind
% submissions: the style file will automatically remove it for you
% unless you've provided the [accepted] option to the icml2023
% package.

% List of affiliations: The first argument should be a (short)
% identifier you will use later to specify author affiliations
% Academic affiliations should list Department, University, City, Region, Country
% Industry affiliations should list Company, City, Region, Country

% You can specify symbols, otherwise they are numbered in order.
% Ideally, you should not use this facility. Affiliations will be numbered
% in order of appearance and this is the preferred way.
\icmlsetsymbol{equal}{*}

\begin{icmlauthorlist}
\icmlauthor{Brandon Morgan}{ou}
\icmlauthor{Dean Hougen}{ou}
%\icmlauthor{}{sch}
%\icmlauthor{}{sch}
\end{icmlauthorlist}

\icmlaffiliation{ou}{School of Computer Science, University of Oklahoma, Oklahoma, USA}

\icmlcorrespondingauthor{Brandon Morgan}{morganscottbrandon@ou.edu}
\icmlcorrespondingauthor{Dean Hougen}{hougen@ou.edu}

% You may provide any keywords that you
% find helpful for describing your paper; these are used to populate
% the "keywords" metadata in the PDF but will not be shown in the document
\icmlkeywords{Loss Function Evolution, Genetic Algorithm, AutoML, MetaLearning, Machine Learning, ICML}

\vskip 0.3in
]

% this must go after the closing bracket ] following \twocolumn[ ...

% This command actually creates the footnote in the first column
% listing the affiliations and the copyright notice.
% The command takes one argument, which is text to display at the start of the footnote.
% The \icmlEqualContribution command is standard text for equal contribution.
% Remove it (just {}) if you do not need this facility.

\printAffiliationsAndNotice{}  % leave blank if no need to mention equal contribution
%\printAffiliationsAndNotice{\icmlEqualContribution} % otherwise use the standard text.

\begin{abstract}
For classification, neural networks typically learn by minimizing cross-entropy, but are evaluated and compared using accuracy. This disparity suggests neural loss function search (NLFS), the search for a drop-in replacement loss function of cross-entropy for neural networks. We apply NLFS to image classifier convolutional neural networks. We propose a new search space for NLFS that encourages more diverse loss functions to be explored, and a surrogate function that accurately transfers to large-scale convolutional neural networks. We search the space using regularized evolution, a mutation-only aging genetic algorithm. After evolution and a proposed loss function elimination protocol, we transferred the final loss functions across multiple architectures, datasets, and image augmentation techniques to assess generalization. In the end, we discovered three new loss functions, called \emph{NeuroLoss1}, \emph{NeuroLoss2}, and \emph{NeuroLoss3} that were able to outperform cross-entropy in terms of a higher mean test accuracy as a simple drop-in replacement loss function across the majority of experiments.
  
\end{abstract}

\section{Introduction}
Convolutional neural networks (CNNs) trained for classification problems on balanced datasets are typically evaluated using accuracy. However, CNNs learn in these scenarios by minimizing cross-entropy, a loss metric from information theory that measures how similar two probability distributions are concerning a random variable. Minimizing cross-entropy reduces the sum of entropies across each discrete class between the actual distribution and the prediction. By minimizing this distributional difference, it is believed that the model’s output will highly correlate with the evaluation metric, accuracy. In practice, this typically holds true; however, a perfect linear mathematical relationship between the two has not been shown. As a result, other possible loss functions could exist that might yield better accuracy than cross-entropy when minimizing. Searching for other candidate loss functions can be referred to as \emph{neural loss function search} (NLFS). 

In this work, the search for a drop-in replacement loss function for cross-entropy on \emph{large-scale CNNs} (those containing millions of parameters or more) for image classification was performed. Each candidate loss function was evaluated on the CIFAR-10 dataset while using \emph{regularized evolution} \citep{AmoebaNet} to explore the search space. The final loss functions were transferred across multiple architectures, datasets, and image augmentation techniques to assess generalization. 

Our summarized contributions are: (1) we propose a new search space for NLFS, allowing for greater exploration of possible loss functions for classification problems; (2) we propose a surrogate function that correlates well to transferable loss functions for large-scale CNNs, even when massive regularization is present; and (3) we discover and present three new loss functions, \emph{NeuroLoss1}, \emph{NeuroLoss2}, and \emph{NeuroLoss3}, that are capable of surpassing cross-entropy as a drop-in replacement across multiple architectures, datasets, and image augmentation techniques.

\section{Related Work}
\label{sec:related}

NLFS is a new but rapidly growing domain \citep{MetaLearning,BAIKAL,LossTaylor,LossObjectDetection,LossLabelNoise,LossDistributionalShift,SoftmaxPersonLoss,AutoLoss,LearnSymbLossMeta}, that originated from the success of neural architecture search (NAS) \citep{NEAT,OGNASNet,NASNet,AmoebaNet,NASSurvey} and AutoML \citep{AUTOML}. NLFS has been explored for various tasks using two primary representations: computational graphs \citep{BAIKAL,LossObjectDetection,SoftmaxPersonLoss,AutoLoss,LearnSymbLossMeta} and Taylor polynomials \citep{LossTaylor,TaylorNPoly}. Here, we focus on computational-graph-based representations.

\citet{BAIKAL} evolved a loss function for classification and regression using genetic programming (GP); the function's coefficients were then fine-tuned using covariance-matrix adaptation evolutionary strategy (CMA-ES). \citet{LossObjectDetection} evolved a loss function specifically for object detection. \citet{LearnSymbLossMeta} found a loss function for classification problems using both GP and a gradient-based local search algorithm. Instead of using CMA-ES to evolve the coefficients of the loss function, %as in \citet{BAIKAL},
%\citet{LearnSymbLossMeta}
they performed meta-learning on the coefficients through gradient descent by comparing with cross-entropy. \citet{AutoLoss} evolved a general loss function for generic tasks in semantic segmentation, object detection, and instance segmentation using GP. \citet{SoftmaxPersonLoss} evolved margin-based softmax loss functions for the specific task of person re-identification using GP. 

In this work, we conduct NLFS for image classification, rather than object detection \citep{LossObjectDetection}, image segmentation \citep{AutoLoss}, or person re-identification \citep{SoftmaxPersonLoss}. %Although previous work exists for applying NLFS to image classification \citep{BAIKAL,LearnSymbLossMeta}, 
We expand on previous image classification NLFS work \citep{BAIKAL,LearnSymbLossMeta} by (1) greatly increasing the allowed operations, (2) using a computational-graph representation rather than grammar-based GP, (3) ensuring our surrogate function transfers effectively to large-scale CNNs with state-of-the-art image regularization techniques, and (4) not biasing our loss functions toward cross-entropy through meta-learning.% \citep{LearnSymbLossMeta}.  

\section{Methodology}
\label{sec:methodology}

In this section, the methodology behind the proposed search space, loss function initialization, loss function integrity check, proposed surrogate function, initial population size de-couplization, loss function elimination protocol, and software implementation details are explained.  

\subsection{Search Space}

The proposed search space for the loss functions uses a computational graph, as with previous work \citep{BAIKAL,LossObjectDetection,SoftmaxPersonLoss,AutoLoss,LearnSymbLossMeta}. However, instead of using the grammar-based representation of those works,
%\cite{BAIKAL,LossObjectDetection,SoftmaxPersonLoss,AutoLoss,LearnSymbLossMeta},
a derivative of the NASNet search space \cite{NASNet} is used.

The NASNet search space optimizes cell-like architectures containing free-floating nodes called \emph{hidden state nodes}. These nodes receive argument connections, perform an operation, and output a value to be received by other nodes. We propose a derivative of this format specifically for loss functions. In addition, our proposed search space expands on previous research by greatly expanding on the possible choices for operations, as previous research use only a few, simple operations \cite{BAIKAL,LossObjectDetection,SoftmaxPersonLoss,AutoLoss,LearnSymbLossMeta}. 

Tree-based grammars are known to suffer two primary issues: bloating \cite{BLOAT} and redundancy \cite{REDUNDANCY}. \emph{Bloating} occurs when the average size of a tree within the population begins to grow uncontrollably. The NASNet search space inherently prevents this as the maximum number of nodes is prescribed prior to evolution, thus limiting the maximum size of a cell. \emph{Redundancy} occurs in tree-based grammars when parts of the tree contain redundant features, which can lead to excessively large trees. Although redundancy is not inherently a problem, it signifies that a particular expression is needed in multiple places, which tree-based grammars must evolve independently. The NASNet search space inherently solves this problem as hidden state nodes are abled to be reused multiple times within the computational graph, thus allowing for redundant features without increasing the number of nodes, thereby speeding up the evolution to acquire such redundant features.  

Unlike \citet{BAIKAL} and \citet{LearnSymbLossMeta}, we do not fine tune the coefficients of the final loss equation. Although fine tuning can greatly increase convergence, this was not performed for two primary reasons. First, to avoid over parameterizing the final loss function on a particular dataset, ensuring simple and transferable loss functions. Second, the computational overhead needed to perform a secondary search is greatly increased when dealing with a continuous domain and an expensive surrogate function. Although meta-learning can be faster to converge than CMA-ES, as it uses a direction gradient to step toward, current meta-learning techniques inherently bias losses towards cross-entropy \citep{MetaLearning,LearnSymbLossMeta}, because they use entropy as the meta-learner when calculating the gradients. In the end, we did not find not fine tuning the coefficients problematic as evidenced by our results.  

The derivative of the NASNet search space proposed in this work has three components: (1) \emph{input connections}, which are analogous to leaf or terminal nodes for tree-based grammars: $y$, $\hat y$, $1$, and $-1$, where $y$ is the target output and $\hat y$ is the model's prediction; (2) \emph{hidden state nodes}, which are free-floating nodes able to use any other hidden node or input connection as an argument; and (3) a specialized \emph{output node}, defined to be the final operation before outputting the loss value. All nodes use either a unary or binary function.

\begin{table}[t]
\caption{The proposed set of unary operations.}
\label{tab:unary}
\vskip 0.1in
\begin{center}
\begin{small}
\begin{tabular}{ll}
\toprule
\begin{sc}
Unary Operations
\end{sc}
 \\
\midrule
$-x$ & $\ln(|x|+\epsilon)$ \\
$e^x$ & $\log_{10}(|x|+\epsilon)$  \\
$\sigma(x)$ & $\frac{d}{dx} \sigma(x)$ \\
$\text{softsign}(x)$ & $\frac{d}{dx}\text{softsign}(x)$  \\
$\text{softplus}(x)$ & $\text{erf}(x)$ \\
$\text{erfc}(x)$ &  $\sin(x)$ \\
$\sinh(x)$ & $\text{arcsinh}(x)$ \\
$\tanh(x)$ & $\frac{d}{dx} \tanh(x)$ \\
$\text{arctanh}(x)$ &  $1/(x+\epsilon)$ \\
$|x|$ & $x^2$ \\
$\text{bessel}_{i0}(x)$ & $\sqrt{x}$ \\
$\text{bessel}_{i1}(x)$ & $\max(x, 0)$ \\
$\text{bessel}_{i1e}(x)$ & $\min(x, 0)$ \\
$\text{bessel}_{i0e}(x)$  & \\
\bottomrule
\end{tabular}
\end{small}
\end{center}
\vskip -0.1in
\end{table}

\begin{table}[t]
\caption{The proposed set of binary operations.}
\label{tab:binary}
\vskip 0.1in
\begin{center}
\begin{small}
\begin{sc}
\begin{tabular}{ll}
\toprule
Binary Operations \\
\midrule
$x_1+x_2$ & $x_1*x_2$ \\
$x_1-x_2$ & $x_1/(x_2+\epsilon)$ \\
$x_1/\sqrt{1+x_2^2}$ & $\max(x_1, x_2)$ \\
$\min(x_1, x_2)$ & \\
\bottomrule
\end{tabular}
\end{sc}
\end{small}
\end{center}
\vskip -0.1in
\end{table}

\begin{figure}[htb]
\vskip 0.1in
\begin{center}
\centerline{\includegraphics[width=0.4\columnwidth]{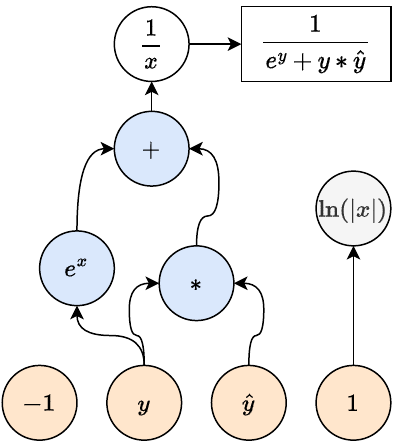}}
\caption{Example loss function %using the proposed search space
with three active (blue) hidden state nodes, one inactive hidden node (grey), and one root node (white). The final loss equation is given next to the root node in the box.} 
\label{fig:ex_loss}
\end{center}
\vskip -0.1in
\end{figure}

%The 27 proposed unary operations are listed in Table \ref{tab:unary}, where $\text{bessel}_{ij}$ is the Bessel function of degree $j$; $e$ denotes exponential; $\sigma(x)=1/(1+e^{-x})$ is sigmoid activation; $\text{softsign}(x)=1/(1+|x|)$ is softsign activation; $\text{softplus}(x)=\ln(1+e^x)$ is softplus activation; $\text{erf}(x)=\frac{2}{\sqrt{\pi}}\int_{0}^{x}e^{-t^2} \mathrm d t$ is the Gauss error function; $\text{erfc}(x)=1-\text{erf}(x)$ is the complementary error function; $\frac{\partial }{\partial x}$ refers the derivative of the respective function; and $\epsilon=10^{-7}$ for numerical stability. The seven proposed binary operations are listed in Table \ref{tab:binary}.

The 27 proposed unary operations are listed in Table~\ref{tab:unary}, where $\text{bessel}_{ij}$ is the Bessel function of degree $j$; $e$ denotes exponential; $\sigma$ is sigmoid activation; $\text{softsign}$ is softsign activation; $\text{softplus}$ is softplus activation; $\text{erf}$ is the Gauss error function; $\text{erfc}$ is the complementary error function; $\frac{d}{dx}$ refers to the derivative of the respective function; and $\epsilon=10^{-7}$ for numerical stability. The seven proposed binary operations are listed in Table~\ref{tab:binary}. All operations were chosen from previous related work in NLFS and available operations in Tensorflow. The goal of increasing the number of possible operators is to create an expansive search space, allowing for loss functions to be expressed in many possible ways. 

\subsection{Loss Function Initialization}
\label{sec:loss_init}

%Initially, each loss function is randomly created using the following procedure.
Each loss function contains four hidden state nodes and one output node. Each node is initialized by sampling its operation uniformly from the set of all 34 operations. Each connection for each hidden state node is sampled uniformly from all input connections and hidden state nodes. The connections for the root node are determined such that at least one connection is from a hidden state node to prevent the output from being a simple binary or unary operation on the input. Figure~\ref{fig:ex_loss} shows an example loss function after initialization.

\subsection{Integrity Check}
\label{sec:integrity}

With the inclusion of many new operations in the proposed search space, and the proposal of free-floating hidden state nodes, the total search space contains a lower bound of $10^{10}$ possible combinations. Although the search space allows for the exploration of many potential good loss functions, it also allows for many degenerate loss functions. To prevent degenerate loss functions from wasting valuable evaluation time, an integrity check was implemented, similar to that by \citet{LossObjectDetection}. If a loss function failed an integrity check, that function was re-initialized using the procedure described in Section \ref{sec:loss_init}. The integrity check contains four components. 

First, the computational graph containing the hidden state nodes for each loss function is checked for cycles. If a cycle is found, the loss function is rejected. In addition, if $y$ or $\hat y$ is not included at least once in the graph, it is also rejected.

\begin{figure}[ht]
\vskip 0.1in
\begin{center}
\centerline{\includegraphics[width=0.6\columnwidth]{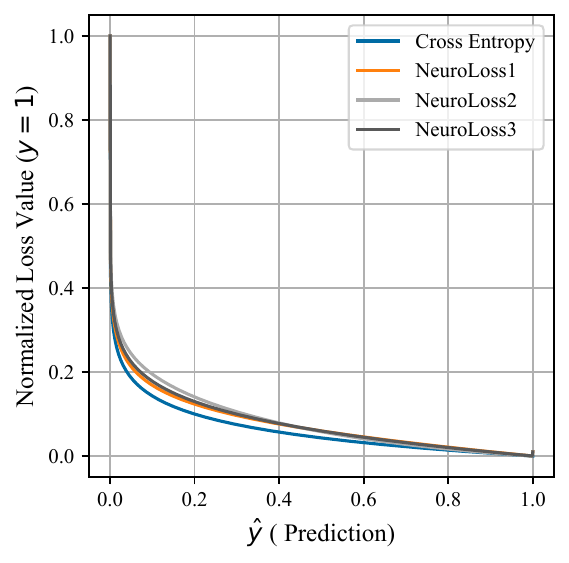}}
\caption{Overview of the binary phenotypes of cross-entropy, NeuroLoss~1, NeuroLoss~2, and NeuroLoss~3.}
\label{fig:phenotype_large}
\end{center}
\vskip -0.1in
\end{figure}

Second, if the first integrity test is passed, the phenotype of the loss function is acquired for a simple binary case. The \emph{phenotype of the loss function} refers to the function value of the loss for given prediction and target arrays, specifically when $y=1$ in this case. The phenotype of cross-entropy, normalized between 0 and 1, using this evaluation is plotted in Figure~\ref{fig:phenotype_large}. As one can see, the loss value for cross-entropy decreases drastically as $\hat y \to 1$, signifying the global minimum, the place where the input-target pairs match. A loss function for a binary case should contain only one global minimum, as there should exist only one solution that minimizes the loss function for all input. Cross-entropy exemplifies this as it is a monotonically decreasing function, as one can see by the figure. To eliminate multi-modal loss landscapes, all loss functions containing more than one global optimum for a binary case are rejected. As a result, the remaining loss functions are either monotonic or roughly parabolic. If the loss function is roughly parabolic, it is rejected if its global optimum is within 1\% of the $\hat y$ value 0.5 to prevent global minimum values near $0.5$, the threshold for accuracy. 

\begin{figure}[ht]
\vskip 0.1in
\begin{center}
\centerline{\includegraphics[width=0.6\columnwidth]{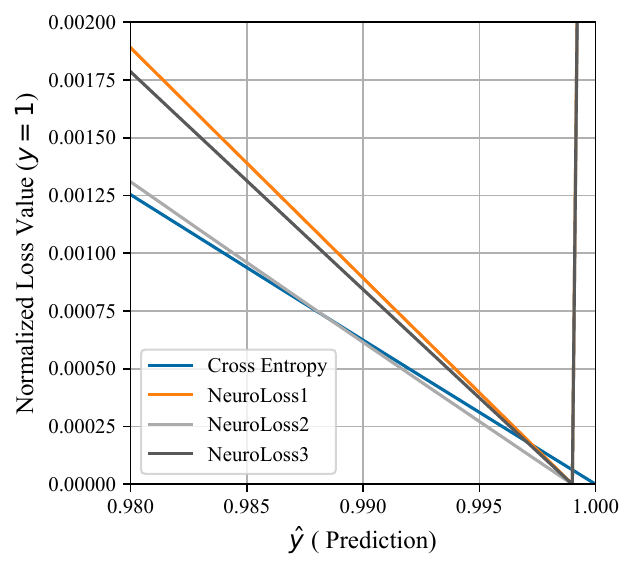}}
\caption{Zoomed in view (around $\hat y\to 1$) of the binary phenotypes of cross-entropy, NeuroLoss~1, NeuroLoss~2, and NeuroLoss~3.}
\label{fig:phenotype_small}
\end{center}
\vskip -0.1in
\end{figure}

Third, the integrity check flips the sign of the loss function if it is either monotonically increasing or an upside-down parabola. This was performed to prevent \emph{inverse learning}---learning an output of 0 for a target output of 1, and an output of 1 for a target output of 0. This simple adjustment inherently speeds up evolution by not requiring loss functions to evolve the proper signage for the root node. 

Fourth, the phenotypes of all existing loss functions within the population, and the current loss function, are normalized between 0 and 1. All normalized phenotypes from the population are looped over and the the Euclidean norm of the difference between each normalized phenotype and the current normalized loss function are calculated. If a single normalized difference was below a threshold of $0.01$ the loss function was rejected for being too similar to another existing loss function. This was performed to allow for ample exploration of non-explored regions of the search space without simply scaling already existing loss functions. However, repeat loss functions were allowed during evolution as long as the copy died off from within the population, a benefit to the aging mechanism (see Section~\ref{sec:ga}).

After passing all four components of the integrity check, each loss function is evaluated using the proposed surrogate function. The final algorithm for the integrity check is detailed in Appendix~\ref{app:algos}.

\subsection{Genetic Algorithm}
\label{sec:ga}
A population-based genetic algorithm (GA) is used to explore the proposed search space. Unlike grammar-based search spaces, our proposed NASNet derivative does not allow for an easy implementation of crossover (recombination), as the hidden state nodes can occur in any order and can connect to any other node. As a result, a mutation-only GA is used. Given the success of regularized evolution \citep{AmoebaNet} on the NASNet search space, we use regularized evolution for our search space. \emph{Regularized evolution} is a mutation-only aging GA that selects a parent through tournament selection from the population. This parent is mutated to obtain a child loss that is evaluated and added back to the population by replacing the oldest. Because training neural networks is highly stochastic, and in order to prevent pre-convergence to local minima and allow for ample exploration, aging is incorporated for better exploration of the search space as each individual has a maximum life span within the algorithm. This is inherently performed in the GA as the new child replaces the oldest.

Mutation was performed by uniformly sampling either the root node or one of the hidden state nodes, then changing either its operation or connection. After mutation, an integrity check (Section~\ref{sec:integrity}) is performed to ensure a successful mutation. If the integrity check fails, mutation is performed again. If 2500 mutations fail in a row, the mutation does not take place, and another iteration of the algorithm is performed using the procedure just discussed. Further information on mutation can be found in Appendix \ref{app:mutation}.

\subsection{Surrogate Function}
\label{sec:surrogate}

It is simple to find specific components of neural networks that perform better than standard versions on a particular dataset or on smaller networks. %In fact, our results will support this claim.
However, it can be very difficult to find components of neural networks that perform better than standard choices across multiple datasets, on models containing millions of parameters, or when massive amounts of regularization are present during training. To address this issue, we choose to compare the final evolved loss functions on EfficientNetV2Small (EffNetV2Small) \citep{EffNetV2}. EffNetV2Small is a state-of-the-art CNN model containing 20.3 million parameters, that was able to achieve a Top-1\% validation accuracy of 83.9\% on ImageNet ILSVRC2012 \citep{ImageNet}. Along with the EffNetV2Small model, we use the same training strategy used by \citet{EffNetV2}: progressive RandAug regularization \citep{RandAug}. \emph{RandAug} is an image augmentation strategy that randomly augments an image by applying $n$ augmentations from a set of possible augmentations, with each augmentation strength capped at a given maximum strength. In \emph{progressive RandAug regularization}, the training strategy is split into four phases, where each hyper-parameter of augmentation is increased throughout training along with image resolution size. We believe that training loss functions using this state-of-the-art model and state-of-the-art regularization technique will accurately compare our final proposed loss functions against cross-entropy for large-scale CNNs. 

Because it is prohibitively expensive to conduct NLFS directly using the full EffNetV2Small training schedule, we use a reduced \emph{surrogate function} during the evolutionary process, as is standard for learning transferable components \citep{NeuralOpt,ACT1,BAIKAL,BNGA}. An issue that can plague learning transferable structures for neural networks is the correlation between the full model and the surrogate model. If there is no correlation between the smaller surrogate model and the full version, then there is no logical reason to use the surrogate. Although it is simple in concept, it can be very difficult to find an accurate surrogate function for learning transferable structures for a particular component. This problem will be further exemplified shortly with our initial findings. Because we care about the relational ranking of loss functions between the surrogate and the full model, Kendall's Tau rank correlation \citep{Kendall} was used instead of Pearson's Correlation. 

As a cheap proxy dataset, all models were trained on the CIFAR-10 dataset \citep{CIFAR}. For promoting generalization, 5,000 of the training images from CIFAR-10 were used as validation. The best validation accuracy was used as the fitness value during evolution, and for calculating rank correlation.

\subsubsection{Chosen Surrogate Function}

In the search for a suitable surrogate function, we looked at three possible models: (1) a three-layer ConvNet, containing 14,000 parameters; (2) a ResNet9 version two (ResNet9V2) \citep{ResNetV2} with nine layers, containing 30,000 parameters; and (3) EffNetV2Small. 

To generate potential loss functions, we performed two preliminary evolutionary runs using the ConvNet and ResNet9 architectures as surrogate functions, both trained on the native dataset image resolution size of 32 for CIFAR-10 without RandAug regularization for 8,000 steps. The best 30 loss functions from each run were then trained using the full EffNetV2Small progressive RandAug training schedule for 64,000 steps. Their final Kendall's Tau rank correlation is shown in the first row of Table~\ref{tab:corr}. Both performed poorly in transferring their final best loss functions to a large-scale model, both achieving a rank correlation of 0.1849, highlighting the issue of finding transferable surrogate functions. 

\begin{table}[htb]
\caption{Kendall's Tau Rank Correlation between ConvNet and ResNet9V2 (ResNet), and EffNetV2Small (EffNet). The first row gives the rank correlation for both ConvNet and ResNet9V2 trained for 8,000 steps without RandAug regularization. The remaining rows give the rank correlation for the architectures trained using RandAug regularization at the given image size for 8,000 steps. }
\label{tab:corr}
\vskip 0.1in
\begin{center}
\begin{small}
\begin{sc}
\begin{tabular}{p{1.2cm} || p{1.3cm} c p{1.3cm} c p{1.3cm} c}
\toprule
Image Size & ConvNet & ResNet & EffNet \\
\midrule
- & 0.1849 & 0.1849 & - \\
\midrule
32 & 0.2352 & 0.1199 &  0.0344\\
48 & 0.1150 & 0.2649 & 0.1448 \\
64 & 0.2768 & 0.1933 & 0.3149 \\
80 & 0.1887 & 0.3498 & 0.2643 \\
96 & 0.2926 & 0.1218 & \textbf{0.5080} \\
112 & 0.2873 & 0.2715 & 0.3420 \\
\bottomrule
\end{tabular}
\end{sc}
\end{small}
\end{center}
\vskip -0.1in
\end{table}

To find a surrogate function with better rank correlation, we took the final 30 loss functions from the ResNet9 evolutionary run (using their associated full scale EffNetV2Small progressive RandAug training schedule validation accuracy), and trained each one on the ConvNet, ResNet9, and EffNetV2Small architectures again, except with progressive RandAug regularization at different image resolution sizes for 8,000 steps. Table~\ref{tab:corr} shows the correlation between each image resolution size and model trained using progressive RandAug regularization and the full scale EffNetV2Small progressive RandAug training schedule. As one can see, the correlations seem almost random, as no apparent relationship can be derived from either the image resolution or architecture. We believe this exemplifies the difficulty of finding a successfully transferable surrogate function. However, a configuration with an acceptable correlation appeared: EffNetV2Small trained using 96$\times$96 image resolution, with a rank correlation of 0.5080. Although EffNetV2Small is more expensive to train than ResNetV2 and ConvNet, we decided to ensure a transferable surrogate function, and selected it as our surrogate function.      

\subsection{Initial Evolutionary Run}

In our initial evolutionary run using the ConvNet architecture as the surrogate function, the best performing loss function was $\frac{1}{n}\sum_i(\text{min}(\text{bessel}_{i0e}(\hat y_i), y_i)+\text{bessel}_{i0e}(\hat y_i))$, which we call the \emph{Bessel Loss}. As a comparison, Table~\ref{tab:convnet_surrogate} shows the mean validation accuracy on the ConvNet surrogate function after 30 independent runs for \emph{Bessel Loss}, cross-entropy, and the final three \emph{NeuroLoss} functions. As one can see, \emph{Bessel Loss} outperformed all \emph{NeuroLoss} functions, and cross-entropy with a higher mean validation accuracy. In addition, a two-sided statistical Welch's $t$-test of the mean difference between \emph{Bessel Loss} and the other loss functions being larger than zero (indicating \emph{Bessel Loss} has a larger mean) was performed. The \emph{Bessel Loss} was shown to have a statistically larger mean \emph{NeuroLoss2} and cross-entropy. However, when training \emph{Bessel Loss} on EffNetV2Small using the full training schedule of progressive RandAug regularization on CIFAR-10, the loss struggles to reach a mean test accuracy of 94.51\%, compared to \emph{NeuroLoss1}'s 95.95\%, \emph{NeuroLoss2}'s 95.91\%, and \emph{NeuroLoss3}'s 95.82\% test accuracy (From Table~\ref{tab:all_model}). This observation reveals that possibly using smaller architectures, with little to no regularization, as a surrogate function for NLFS may not be beneficial as their performance may not correlate well to large-scale CNNs with regularization.

\begin{table}[htb]
\caption{Results for Bessel Loss, NeuroLoss1, NeuroLoss4, and cross-entropy using the ConvNet surrogate function. (NL$_1$ for NeuroLoss~1, etc.)
The mean test accuracy, and standard deviation, across 30 independent runs are reported for each loss function along with the $p$-value for a one sided statistical $t$-test for mean accuracy being larger than cross-entropy. The best mean test accuracy is highlighted in bold.}
\label{tab:convnet_surrogate}
\vskip 0.1in
\begin{center}
\begin{small}
\begin{sc}
\begin{tabular}{p{1.75cm} l p{1.9cm} c p{1.6cm} r}
\multicolumn{3} {  c  }{ConvNet Surrogate Function}\\
\toprule
Loss & Validation Accuracy & $p$-value\\
\midrule
BesselLoss & \textbf{58.839}$\pm$0.572 & - \\
NL$_1$ & 58.711$\pm$0.455 & $p = 0.171$ \\
NL$_2$ & 58.579$\pm$0.612 & $p=0.047$\\
NL$_3$ &  58.660$\pm$0.557 &  $p=0.112$\\
CE & 58.474$\pm$0.569 & $p=0.008$\\
\bottomrule
\end{tabular}
\end{sc}
\end{small}
\end{center}
\vskip -0.1in
\end{table}

\subsection{Early Stopping}

To avoid wasting time in training degenerate loss functions, an early stopping method was implemented. Before training on EffNetV2Small 96$\times$96 resolution, a ConvNet was trained on the images, without regularization, for 800 steps. If the training accuracy was below a hand-tuned threshold of 37\%, the loss function was rejected and mutated again. If the training accuracy was above this threshold, the loss was trained on the surrogate function, where a secondary early stopping method was implemented. This secondary early stopping method was different from the first. If the training accuracy had not reached a hand-tuned threshold of 25\% after 3,000 steps, training was stopped; however, the best validation accuracy was returned as the fitness score to the genetic algorithm. 

\subsection{Initial Population}

In our initial shortened evolution we noted that the majority of loss functions for the initial population of the algorithm were degenerate in that the first early stopping mechanism implemented within the surrogate function stopped training due to poor learning. Allowing for degenerate models within the initial population caused three primary issues. First, it created weak competition within the population, as working loss functions would have less competition for mutation. Second, it allowed for early pre-convergence, as only the smaller set of working functions were chosen for mutation. Third, for tournaments containing only degenerate losses, the child loss of the degenerate parent would inherently have higher probability of being degenerate as well. We believe this allowance will degrade evolution quality in the long run. As a possible solution, we decided to decouple the initial population size from the population size. In doing so, we increased the initial population size to allow for more diversity and to allow for more working loss functions. The best from the initial population were then taken to form the first generation, ensuring working initial loss functions.

\subsection{Loss Function elimination Protocol}

Due to the disparity of correlation between the surrogate function and full EffNetV2Small training, we decided to spend the majority of our computational resources on training for transferability rather than searching for loss functions, as our integrity check and early stopping mechanisms eliminated most degenerate solutions. To do so, we implemented a loss function elimination protocol that progressively eliminated candidate loss functions, after evolution, until a final set of winners remained by progressively training them on the different phases of RandAug regularization for EffNetV2Small. The best 150 loss functions discovered over the entire course of evolution were trained on EffNetV2Small for phase~1 (16,000 steps) of RandAug progressive regularization. The best 50 from that set were then trained again, from scratch, for both phase~1 and 2 (32,000 steps) of RandAug regularization. The best 25 from that set were then trained again until phase~3 (48,000 steps) of RandAug regularization. Finally, the best 10 were taken and trained for the full RandAug progressive regularization training strategy (64,000 steps). Note, all loss functions in our proposed loss function elimination protocol were compared using the validation accuracy for CIFAR-10. 

\subsection{Implementation}

For the GA, an initial population size of 1000 was used to allow for ample initial diversity and working loss functions. The best 100 were then taken to form the first generation. A tournament size of 20 was used for selection. The GA was allowed to run for a time frame of 100 hours on an NVIDIA A100 GPU. Each individual within the population was trained using the proposed surrogate function, EffNetV2Small 96$\times$96 resolution for 8,000 steps of Phase~1 of RandAug progressive regularization. Due to computational constraints, the initial population was not trained using this surrogate function, but was instead trained using the ConvNet architecture without RandAug regularization. We did not find this problematic as the purpose of the enlarged initial population was to find workable loss functions in a short amount of time. 

\section{Results and Discussion}
\label{sec:results}

In this section, the results from the evolutionary run are detailed, along with the final best losses and their evaluation across multiple datasets and architectures. 

\subsection{Evolution}

\begin{figure}[htb]
\vskip 0.1in
\begin{center}
\centerline{\includegraphics[width=0.7\columnwidth]{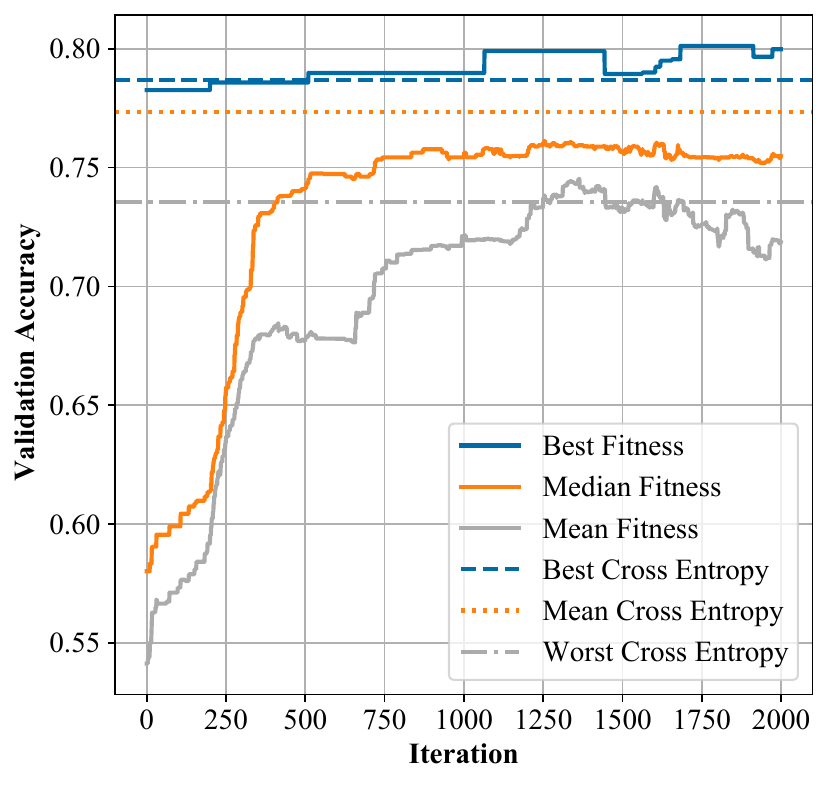}}
\caption{Results for the evolutionary algorithm evaluated using the proposed surrogate function.} 
\label{fig:evolution}
\end{center}
\vskip -0.1in
\end{figure}

After 100 hours of evolution on the CIFAR-10 dataset, a total of 2,000 iterations had completed, 29\% of which did not fail the surrogate function. The best, mean, and median validation scores over time are shown in Figure \ref{fig:evolution}. In addition, the max, mean, and min cross-entropy scores across 15 independent training sessions for the surrogate function are plotted for comparison. %As one can see from the figure,
The algorithm was able to evolve 23 loss functions that were able to surpass the best cross-entropy accuracy on the surrogate function, while 119 loss functions were able to surpass the mean cross-entropy accuracy. 

\subsection{Best Losses}

The final three remaining loss functions, which we call the \emph{NeuroLoss} functions, after running our loss function elimination protocol are listed in Table~\ref{tab:final_losses}. As one can see, the final three functions are very similar as all of them follow the form $\log(y, \hat y)+f(y, \hat y)$, where $\log(y, \hat y)$ was either $\ln$ or $\log_{10}$ and $f(y, \hat y)$ was some function. In addition, one can also see the repeated structure of $\hat y/(y+\epsilon)$ that is present twice in almost all functions, revealing the power of re-usable hidden state nodes in our proposed search space. It seems that the final loss functions heavily relied on scaling the prediction $\hat y$ by some value, mainly $\sqrt{1+(\hat y/(y+\epsilon))^2}$ inside the logarithmic operation. Honorable mentions after running the proposed loss function elimination protocol are found in Table~\ref{tab:loss_honarable}.

\begin{table}[t]
\caption{The final three evolved loss functions (NL$_1$ for NeuroLoss~1, etc.), along with cross-entropy (CE) for comparison.}
\label{tab:final_losses}
\vskip 0.1in
\begin{center}
\begin{small}
\setlength{\tabcolsep}{1pt}
\begin{sc}
\begin{tabular}{cc}
\toprule
& Loss Equation \\
\midrule
NL$_1$ & \thinmuskip=0mu \medmuskip=0mu \thickmuskip=0mu {\scriptsize$-\frac{1}{n}\sum_i\Bigl(\ln\Bigl(\Bigl|\frac{\hat y_i}{\sqrt{1+(\hat y_i / (y_i+\epsilon))^2}}\Bigr|+\epsilon\Bigr)+\hat y_i \frac{\hat y_i}{\sqrt{1+(\hat y_i / (y_i+\epsilon))^2}}\Bigr)$} \\
NL$_2$ & \thinmuskip=0mu \medmuskip=0mu \thickmuskip=0mu \scriptsize $-\frac{1}{n}\sum_i\Bigl(\log_{10}\Bigl(\Bigl|\frac{\hat y_i}{\sqrt{1+(\hat y_i / (y_i+\epsilon))^2}}\Bigr|+\epsilon\Bigr)+\sin\Bigl(\frac{\hat y_i}{\sqrt{1+(\hat y_i / (y_i+\epsilon))^2}}\Bigr)\Bigr)$ \\
NL$_3$ & \thinmuskip=0mu \medmuskip=0mu \thickmuskip=0mu \scriptsize$-\frac{1}{n}\sum_i\Bigl(\ln\Bigl(\Bigl|\frac{\hat y_i}{\sqrt{1+(\hat y_i / (y_i+\epsilon))^2}}\Bigr|+\epsilon\Bigr)+\min(y_i, \hat y_i / (y_i+\epsilon))\Bigr)$\\[1ex]
CE & $-\frac{1}{n}\sum_i y_i\ln\Bigl(|\hat y_i|+\epsilon\Bigr)$ \\
\bottomrule
\end{tabular}
\end{sc}
\end{small}
\end{center}
\vskip -0.1in
\end{table}

\begin{table}[htb]
\caption{Honorable Mentions}
\label{tab:loss_honarable}
\vskip 0.15in
\begin{center}
\begin{small}
\setlength{\tabcolsep}{1pt}
\begin{tabular}{c}
\toprule
\textsc{Loss Equation} \\
\midrule

\thinmuskip=0mu \medmuskip=0mu \thickmuskip=0mu \scriptsize$-\frac{1}{n}\sum_i\Bigl(\ln\Bigl(\Bigl|\frac{\hat y_i}{\sqrt{1+(y_i)^2}}\Bigr|+\epsilon\Bigr)\Bigr)\big/\Bigl(\sqrt{1+(\max(y_i, (\hat y_i / (y+_i\epsilon)))^2)}\Bigr)$ \\

\thinmuskip=0mu \medmuskip=0mu \thickmuskip=0mu \scriptsize$-\frac{1}{n}\sum_i\max\Bigl(\log(|\frac{\hat y_i}{\sqrt{1+(\hat y_i / (y_i+\epsilon))^2}}|+\epsilon), \text{arcsinh}(\frac{\hat y_i}{\sqrt{1+(\hat y_i / (y_i+\epsilon))^2}})\Bigr)$
\\
\thinmuskip=0mu \medmuskip=0mu \thickmuskip=0mu \scriptsize$-\frac{1}{n}\sum_i\Bigl(\log\Bigl(|\frac{\hat y_i}{\sqrt{1+(\hat y_i / (y_i+\epsilon))^2}}|+\epsilon\Bigr)+(\frac{\hat y_i}{\sqrt{1+(\hat y_i / (y_i+\epsilon))^2}})^2\Bigr)$
\\
\thinmuskip=0mu \medmuskip=0mu \thickmuskip=0mu \scriptsize$-\frac{1}{n}\sum_i\Bigl(\ln\Bigl(|\frac{\hat y_i}{\sqrt{1+(\hat y_i / (y_i+\epsilon))^2}}|+\epsilon\Bigr)+\sinh(\frac{\hat y_i}{\sqrt{1+(\hat y_i / (y_i+\epsilon))^2}})\Bigr)$
\\
\thinmuskip=0mu \medmuskip=0mu \thickmuskip=0mu \scriptsize$-\frac{1}{n}\sum_i\Bigl(\ln\Bigl(|\frac{\hat y_i}{\sqrt{1+(\hat y_i / (y_i+\epsilon))^2}}|+\epsilon\Bigr)+\text{erf}\Bigl(\hat y_i / (y_i+\epsilon))\Bigr)\Bigr)$
\\
\thinmuskip=0mu \medmuskip=0mu \thickmuskip=0mu \scriptsize$-\frac{1}{n}\sum_i\Bigl(\log\Bigl(|\frac{\hat y_i}{\sqrt{1+(\hat y_i / (y_i+\epsilon))^2}}|+\epsilon\Bigr)+\tanh\Bigl(\hat y_i / (y_i+\epsilon)\Bigr)\Bigr)$
\\
\thinmuskip=0mu \medmuskip=0mu \thickmuskip=0mu \scriptsize$-\frac{1}{n}\sum_i\frac{\ln\Bigl(|y_i*\hat y_i|+\epsilon\Bigr)}{\sqrt{1+(\hat y_i / (y_i+\epsilon))^2}}$ 
\\
\bottomrule
\end{tabular}
\end{small}
\end{center}
\vskip -0.1in
\end{table}

\subsection{Evaluation}

To evaluate the generalizability of the final three loss functions, they were transferred across multiple data sets, architectures, image augmentation techniques, and learning rate schedules and optimizers. Six primary datasets were used in this work, CIFAR-10, CIFAR-100, Stanford-CARS196, Oxford-Flowers102, Caltech101, and ImageNet. The meta-information for each of these datasets are available in Table~\ref{tab:datasets}.

\begin{table}[htb]
\caption{The number of training and evaluation images, number of classes, and mean image resolution for the datasets used in this work.}
\label{tab:datasets}
\vskip 0.1in
\begin{center}
\begin{small}
\begin{sc}
\begin{tabular}{ p{1.5cm}p{1.0cm}p{1.0cm}p{1.2cm}p{1.4cm}}
\toprule
Dataset & Train & Test & Classes & Size \\
\midrule
CIFAR10 & 50,000 & 10,000 & 10 & 32$\times$32 \\
CIFAR100 & 50,000 & 10,000 & 100 & 32$\times$32 \\
CARS196 & 8,144 & 8,041 & 196 & 482$\times$700 \\
Flowers102 & 2,040 & 6,149 & 102 & 546$\times$624 \\
Caltech101 & 3,060 & 6,084 & 102 & 386$\times$468 \\
ImageNet & 1.28M & 50,000 & 1000 & 469$\times$387 \\
\bottomrule
\end{tabular}
\end{sc}
\end{small}
\end{center}
\vskip -0.1in
\end{table}

Other than EffNetV2Small, the final loss functions were trained on three other state-of-the art model architectures to assess architecture transferability: WideResNet (depth=28, width=10, 36 million parameters) \citep{WideResNet}, SEResNeXt (4$\times$64d, 20 million parameters) \citep{SERESNET}, and DavidNet (custom ResNet9, 6 million parameters, Appendix~\ref{app:train_strat}). Three different image augmentation techniques were used: progressive RandAug regularization for EffNetV2Small; simple cutout with random cropping and horizontal flipping for WideResNet, SEResNeXt, and DavidNet; and CutMix \citep{CutMix} for DavidNet. CutMix was used as it mixes label distribution, testing for whether or not the final loss functions could handle target values not equal to 0 or 1. To assess dataset transferability, the final loss functions were trained using the previous architectures and image augmentation techniques on both CIFAR-10 and CIFAR-100. Lastly, the final loss functions were transferred to three large image resolution benchmark dataset: Cars196, Flowers102, and Caltech101. Using EffNetV2Small, the final loss functions were trained both from scratch and fine-tuning (using the official published ImageNet weights), on the large image resolution datasets. Details on the training strategies used for each architecture can be found in Appendix~\ref{app:train_strat}.

\subsection{Results}

\begin{table*}[h!]
\caption{Results for DavidNet using resize, cutout, and horizontal flipping, and CutMix for both CIFAR-10 and CIFAR-100. The mean test accuracy and standard deviation across 30 independent runs is reported for each loss function along with the $p$-value for a Welch's two-sided statistical $t$-test for mean difference between the \emph{NeuroLoss} function and cross-entropy being larger than zero (indicating the function has a larger mean). The best mean test accuracy for each dataset is highlighted in bold. }
\label{tab:davidNet}
\vskip 0.1in
\begin{center}
\begin{small}
\begin{sc}
\begin{tabular}{p{0.75cm} l p{1.9cm} c p{1.7cm} r}
&
\multicolumn{2} {  c  }{CIFAR-10 (cutout)} &
\multicolumn{2} {  c  }{CIFAR-10 (cutmix)}\\
\toprule
Loss & Test Accuracy & $p$-value & Test Accuracy & $p$-value\\
\midrule
NL$_1$ & \textbf{94.042}$\pm$0.126 & $p<0.001$ & \textbf{91.621}$\pm$0.184 & $p<0.001$\\  
NL$_2$ & 93.787$\pm$0.138 & $p=0.298$ & 91.409$\pm$0.178 & $p=0.581$\\
NL$_3$ & 94.006$\pm$0.134 & $p<0.001$ & 91.601$\pm$0.169 & $p< 0.001$\\
CE & 93.761$\pm$0.134 & - & 91.419$\pm$0.197 & -\\
& & \\
&
\multicolumn{2} {  c  }{CIFAR-100 (cutout)} &
\multicolumn{2} {  c  }{CIFAR-100 (cutmix)}\\
\toprule
Loss & Test Accuracy & $p$-value & Test Accuracy & $p$-value \\
\midrule
NL$_1$ & \textbf{74.959}$\pm$0.212 & $p=0.003$ & 70.877$\pm$0.270 & $p<0.001$\\  
NL$_2$ & 74.814$\pm$0.254 & $p=0.421$ & 70.736$\pm$0.244 & $p<0.001$\\ 
NL$_3$ & 74.834$\pm$0.263 & $p=0.302$ & \textbf{71.176}$\pm$0.288& $p<0.001$\\ 
CE & 74.802$\pm$0.208 & - & 69.534$\pm$0.228 & -\\ 
\bottomrule
\end{tabular}
\end{sc}
\end{small}
\end{center}
\vskip -0.1in
\end{table*}

\begin{table}[h!]
\caption{Results for EffNetV2Small using progressive RandAug regularization, WideResNet (depth=10, width=28) and SEResNeXt (4x64d) using resize, cutout, and horizontal flipping augmentation for both CIFAR-10 and CIFAR-100. The mean test accuracy and standard deviation across 5 independent runs is reported for each loss function. The best mean test accuracy for each combination of dataset and architecture is highlighted in bold.}
\label{tab:all_model}
\vskip 0.1in
\begin{center}
\begin{small}
\begin{sc}
\begin{tabular}{p{.75cm}p{1.75cm}p{1.75cm}p{1.75cm}}
\multicolumn{4} {  c  }{CIFAR-10}\\
\toprule
Loss & EffNetV2S & WideResNet & SEResNeXt \\
\midrule
Bessel & 94.510$\pm$0.115 \\
NL$_1$ & \textbf{95.952}$\pm$0.106 & \textbf{95.860}$\pm$0.142 & \textbf{95.830}$\pm$0.123 \\
NL$_2$ & 95.906$\pm$0.050 & 95.836$\pm$0.052 & 95.604$\pm$0.032\\
NL$_3$ & 95.816$\pm$0.060 & 95.764$\pm$0.050 & 95.788$\pm$0.078\\
CE & 95.826$\pm$0.118 & 95.670$\pm$0.156 & 95.222$\pm$0.063  \\
& & \\
\multicolumn{4} { c }{CIFAR-100}\\
\toprule
Loss & EffNetV2S & WideResNet & SEResNeXt \\
\midrule
NL$_1$ & 78.172$\pm$0.249 & 79.848$\pm$0.061 & 79.964$\pm$0.167\\
NL$_2$ & 77.926$\pm$0.189 & 79.628$\pm$0.242 &  80.152$\pm$0.263\\
NL$_3$ & \textbf{78.230}$\pm$0.154 & \textbf{80.038}$\pm$0.271 & 79.804$\pm$0.263\\
CE & 77.130$\pm$0.228 & 79.640$\pm$0.213 & \textbf{80.526}$\pm$0.228\\
\bottomrule
\end{tabular}
\end{sc}
\end{small}
\end{center}
\vskip -0.1in
\end{table}

\begin{table}[!ht]
\caption{Results for EffNetV2Small using progressive RandAug regularization when training from scratch for Stanford-CARS196, Oxford-Flowers102, and Caltech101. The mean test accuracy and standard deviation across 5 independent runs is reported for each loss function. The best mean test accuracy for each dataset is highlighted in bold. CE$_{0.10}$ refers to cross-entropy with a label smoothing value of $0.10$.}
\label{tab:effnet_three_datasets_scratch}
\vskip 0.1in
\begin{center}
\begin{small}
\begin{sc}
\begin{tabular}{p{.75cm}p{1.75cm}p{1.75cm}p{1.75cm}}
\multicolumn{4} {  c  }{EffNetV2Small (From Scratch)}\\
\toprule
Loss & CARS196 & Flowers102 & Caltech101\\
\midrule
NL$_1$ & 85.457$\pm$0.244 & \textbf{88.584}$\pm$0.508 & \textbf{85.148}$\pm$0.210\\
NL$_2$ & 85.698$\pm$0.361 & 87.380$\pm$0.461 & 84.234$\pm$0.341 \\
NL$_3$ & \textbf{86.439}$\pm$0.418 & 88.408$\pm$0.440 &  84.822$\pm$0.285\\
CE & 82.729$\pm$0.285 & 88.154$\pm$0.299 & 84.116$\pm$0.119\\
CE$_{0.10}$ & 84.674$\pm$0.254& 87.819$\pm$0.228& 84.499$\pm$0.332 \\
\bottomrule
\end{tabular}
\end{sc}
\end{small}
\end{center}
\vskip -0.1in
\end{table}

\begin{table}[h!]
\caption{Results for EffNetV2Small using progressive RandAug regularization when fine-tuning for Stanford-CARS196, Oxford-Flowers102, and Caltech101. The mean test accuracy and standard deviation across 5 independent runs is reported for each loss function. The best mean test accuracy for each dataset is highlighted in bold. CE$_{0.10}$ refers to cross-entropy with a label smoothing value of $0.10$.}
\label{tab:effnet_three_datasets_fine}
\vskip 0.1in
\begin{center}
\begin{small}
\begin{sc}
\begin{tabular}{p{.75cm}p{1.75cm}p{1.75cm}p{1.75cm}}
\multicolumn{4} {  c  }{EffNetV2Small (Fine-Tune)}\\
\toprule
Loss & CARS196 & Flowers102 & Caltech101\\
\midrule
NL$_1$ & 91.635$\pm$0.109&97.313$\pm$0.095& 90.556$\pm$0.279\\
NL$_2$ & 91.660$\pm$0.053& \textbf{97.421}$\pm$0.139& \textbf{90.565}$\pm$0.428\\
NL$_3$ & 91.635$\pm$0.091&97.287$\pm$0.138& 90.302$\pm$0.295\\
CE & \textbf{91.725}$\pm$0.233& 97.190$\pm$0.188& 90.217$\pm$0.250\\
CE$_{0.10}$ & 91.496$\pm$0.122& 97.320$\pm$0.277 & 90.434$\pm$0.141 \\
\bottomrule
\end{tabular}
\end{sc}
\end{small}
\end{center}
\vskip -0.1in
\end{table}

Results for DavidNet on CIFAR are given in Table~\ref{tab:davidNet}. Results for EffNetV2Small, WideResNet, and SEResNeXt on CIFAR are given in Table~\ref{tab:all_model}. Results for EffNetV2Small on Cars196, Flowers102, and Caltech101 from scratch are given in Table~\ref{tab:effnet_three_datasets_scratch}, while the results for fine-tuning are given in Table~\ref{tab:effnet_three_datasets_fine}. As seen in Table~\ref{tab:davidNet}, multiple loss functions achieved significant $p$-values for having a larger mean test accuracy than cross-entropy across both datasets and image augmentation techniques. However, only one loss, NeuroLoss~1, obtained a significant value for all datasets and augmentation techniques, revealing that it can outperform and increase generalization better than cross-entropy in a computationally stringent environment for DavidNet.

From Tables \ref{tab:davidNet} and \ref{tab:all_model}, only once was the mean test accuracy of cross-entropy better than the best loss function (SEResNeXt for CIFAR-100). Two loss functions dominated the results, NeuroLoss~1 and NeuroLoss~3. In terms of mean test value, NeuroLoss~1 and NeuroLoss~3 surpassed cross-entropy across all datasets and architectures (except NeuroLoss~3 for EffNetV2Small for CIFAR-10, and both losses for SEResNeXt for CIFAR-100). However, NeuroLoss~1 was able to surpass NeuroLoss~3 for all architectures trained upon CIFAR-10, but NeuroLoss~3 was able to surpass Loss~1 and cross-entropy for all architectures trained on CIFAR-100 (except DavidNet using cutout and SEResNeXt). 

From Tables Table~\ref{tab:effnet_three_datasets_scratch}, both NeuroLoss~1 or NeuroLoss~3 outperformed cross-entropy, even with a label smoothing constant of $\alpha=0.10$, across all datasets. However, when changing from training froms scratch to fine-tuning using cross-entropy biased initial weights, Table~\ref{tab:effnet_three_datasets_fine} reports that NeuroLoss~2 outperforms all other NeuroLoss functions and cross-entropy, both with and without label smoothing, on two of the three datasets. With NeuroLoss~2 trailing the other two NeuroLoss functions when training from scratch on Flowers102 and Caltech101, but outperforming both when fine-tuning could indicate that some loss functions might be better inclined to fine-tuning than others.

\subsection{What Makes \emph{NeuroLoss} Successful?}
\label{app:neuro_succ}
Figures~\ref{fig:phenotype_large} and \ref{fig:phenotype_small} depict the phenotypes of all NeuroLoss functions and cross-entropy in a binary example. From both of these figures, there are two noticeable differences. First, all NeuroLoss functions have a less steep slope compared to cross-entropy, resulting in higher normalized loss values. From Figure~\ref{fig:phenotype_large}, NeuroLoss~2 has the highest normalized loss values compared to all other functions when $\hat y < 0.4$; however, Figure~\ref{fig:phenotype_small} indicates that NeuroLoss~2 dips below the other NeuroLoss functions and even cross-entropy as $\hat y \to 1$. NeuroLoss~1 and NeuroLoss~2 seem to hug each other for both small and large $\hat y$ values. Perhaps this more gentle slope could encourage more smoothness in the loss landscape than cross-entropy. However, further investigation and study must be conducted to verify that hypothesis. Second, the global minimum for all NeuroLoss functions do not occur at $\hat y=1$, as with cross-entropy, but instead at $\hat y =0.99998$. This small level of implicit regularization can be replicated in cross-entropy with an extremely small label smoothing constant of $\alpha=0.00003$. Although the global minimum can be shifted in cross-entropy via label smoothing to match that of the NeuroLoss functions, the slope cannot be changed to equal them. The exact reason for the \emph{NeuroLoss} functions success is unknown, but two possible reasons arise. The implicit learning of an extremely small label smoothing constant and a higher normalized loss values than cross-entropy over the majority of $\hat y$ values. 

\section{Conclusions}
\label{sec:conclusions}

In this work, we expanded on previous research in NLFS by proposing a new search space containing more unary and binary operations, along with a new computational graph representation that eliminated some of the downsides associated with GP. In addition, we stressed the importance of ensuring transferable surrogate functions to ensure good rank correlation. After running our evolutionary algorithm and loss function elimination protocol, three loss functions highly competitive with cross-entropy were discovered through pure evolution. The best loss functions, termed \emph{NeuroLoss}~1, \emph{NeuroLoss}~2, and \emph{NeuroLoss}~3, were able to surpass cross-entropy across multiple architectures, datasets, and image augmentation techniques. The proposed search space, integrity check, and elimination protocol can be used for future NFLS endeavors, such as loss functions for image anomaly detection, imbalanced image datasets, or traditional based machine learning algorithms, such as logistic regression and decision trees.

%%
%% The acknowledgments section is defined using the "acks" environment
%% (and NOT an unnumbered section). This ensures the proper
%% identification of the section in the article metadata, and the
%% consistent spelling of the heading.

%%
%% The next two lines define the bibliography style to be used, and
%% the bibliography file.
\bibliography{main}

\begin{thebibliography}{36}
\providecommand{\natexlab}[1]{#1}
\providecommand{\url}[1]{\texttt{#1}}
\expandafter\ifx\csname urlstyle\endcsname\relax
  \providecommand{\doi}[1]{doi: #1}\else
  \providecommand{\doi}{doi: \begingroup \urlstyle{rm}\Url}\fi

\bibitem[Abadi et~al.(2015)Abadi, Agarwal, Barham, Brevdo, Chen, Citro, Corrado, Davis, Dean, Devin, Ghemawat, Goodfellow, Harp, Irving, Isard, Jia, Jozefowicz, Kaiser, Kudlur, Levenberg, Man\'{e}, Monga, Moore, Murray, Olah, Schuster, Shlens, Steiner, Sutskever, Talwar, Tucker, Vanhoucke, Vasudevan, Vi\'{e}gas, Vinyals, Warden, Wattenberg, Wicke, Yu, and Zheng]{Tensorflow}
Abadi, M., Agarwal, A., Barham, P., Brevdo, E., Chen, Z., Citro, C., Corrado, G.~S., Davis, A., Dean, J., Devin, M., Ghemawat, S., Goodfellow, I., Harp, A., Irving, G., Isard, M., Jia, Y., Jozefowicz, R., Kaiser, L., Kudlur, M., Levenberg, J., Man\'{e}, D., Monga, R., Moore, S., Murray, D., Olah, C., Schuster, M., Shlens, J., Steiner, B., Sutskever, I., Talwar, K., Tucker, P., Vanhoucke, V., Vasudevan, V., Vi\'{e}gas, F., Vinyals, O., Warden, P., Wattenberg, M., Wicke, M., Yu, Y., and Zheng, X.
\newblock {TensorFlow}: Large-scale machine learning on heterogeneous systems, 2015.
\newblock URL \url{https://www.tensorflow.org/}.
\newblock Software available from tensorflow.org.

\bibitem[Anonymous(2023)]{CodeNeural}
Anonymous.
\newblock Neural loss function evolution for large scale image classifier convolutional neural networks.
\newblock \url{https://github.com.invalid/Anon/NeuralLossFunctionEvolution}, 2023.

\bibitem[Bello et~al.(2017)Bello, Zoph, Vasudevan, and Le]{NeuralOpt}
Bello, I., Zoph, B., Vasudevan, V., and Le, Q.~V.
\newblock Neural optimizer search with reinforcement learning.
\newblock In Precup, D. and Teh, Y.~W. (eds.), \emph{Proceedings of the 34th International Conference on Machine Learning}, volume~70 of \emph{Proceedings of Machine Learning Research}, pp.\  459--468. PMLR, 06--11 Aug 2017.
\newblock URL \url{https://proceedings.mlr.press/v70/bello17a.html}.

\bibitem[Bingham et~al.(2020)Bingham, Macke, and Miikkulainen]{ACT1}
Bingham, G., Macke, W., and Miikkulainen, R.
\newblock Evolutionary optimization of deep learning activation functions.
\newblock In \emph{Proceedings of the 2020 Genetic and Evolutionary Computation Conference}, GECCO '20, pp.\  289–296, New York, NY, USA, 2020. Association for Computing Machinery.
\newblock ISBN 9781450371285.
\newblock \doi{10.1145/3377930.3389841}.
\newblock URL \url{https://doi.org/10.1145/3377930.3389841}.

\bibitem[Chebotar et~al.(2019)Chebotar, Molchanov, Bechtle, Righetti, Meier, and Sukhatme]{MetaLearning}
Chebotar, Y., Molchanov, A., Bechtle, S., Righetti, L., Meier, F., and Sukhatme, G.~S.
\newblock Meta learning via learned loss.
\newblock In \emph{International Conference on Pattern Recognition (ICPR)}, pp.\  4161--4168, 2019.

\bibitem[Coleman et~al.(2019)Coleman, Kang, Narayanan, Nardi, Zhao, Zhang, Bailis, Olukotun, R\'{e}, and Zaharia]{DawnBench}
Coleman, C., Kang, D., Narayanan, D., Nardi, L., Zhao, T., Zhang, J., Bailis, P., Olukotun, K., R\'{e}, C., and Zaharia, M.
\newblock Analysis of {DAWNBench}, a time-to-accuracy machine learning performance benchmark.
\newblock \emph{SIGOPS Oper. Syst. Rev.}, 53\penalty0 (1):\penalty0 14–25, July 2019.
\newblock ISSN 0163-5980.
\newblock \doi{10.1145/3352020.3352024}.
\newblock URL \url{https://doi.org/10.1145/3352020.3352024}.

\bibitem[Cubuk et~al.(2020)Cubuk, Zoph, Shlens, and Le]{RandAug}
Cubuk, E.~D., Zoph, B., Shlens, J., and Le, Q.~V.
\newblock Randaugment: Practical automated data augmentation with a reduced search space.
\newblock In \emph{Proceedings of the IEEE/CVF Conference on Computer Vision and Pattern Recognition Workshops}, pp.\  702--703, 2020.

\bibitem[Deng et~al.(2009)Deng, Dong, Socher, Li, Li, and Fei-Fei]{ImageNet}
Deng, J., Dong, W., Socher, R., Li, L.-J., Li, K., and Fei-Fei, L.
\newblock Imagenet: A large-scale hierarchical image database.
\newblock In \emph{2009 IEEE Conference on Computer Vision and Pattern Recognition}, pp.\  248--255, 2009.
\newblock \doi{10.1109/CVPR.2009.5206848}.

\bibitem[Elsken et~al.(2019)Elsken, Metzen, and Hutter]{NASSurvey}
Elsken, T., Metzen, J.~H., and Hutter, F.
\newblock Neural architecture search: A survey.
\newblock \emph{The Journal of Machine Learning Research}, 20\penalty0 (1):\penalty0 1997--2017, 2019.

\bibitem[Gao et~al.(2021)Gao, Gouk, and Hospedales]{LossLabelNoise}
Gao, B., Gouk, H., and Hospedales, T.~M.
\newblock Searching for robustness: Loss learning for noisy classification tasks.
\newblock In \emph{Proceedings of the IEEE/CVF International Conference on Computer Vision (ICCV)}, pp.\  6670--6679, October 2021.

\bibitem[Gao et~al.(2022)Gao, Gouk, Yang, and Hospedales]{LossDistributionalShift}
Gao, B., Gouk, H., Yang, Y., and Hospedales, T.
\newblock Loss function learning for domain generalization by implicit gradient.
\newblock In Chaudhuri, K., Jegelka, S., Song, L., Szepesvari, C., Niu, G., and Sabato, S. (eds.), \emph{Proceedings of the 39th International Conference on Machine Learning}, volume 162 of \emph{Proceedings of Machine Learning Research}, pp.\  7002--7016. PMLR, 17--23 Jul 2022.
\newblock URL \url{https://proceedings.mlr.press/v162/gao22b.html}.

\bibitem[Gonzalez \& Miikkulainen(2020)Gonzalez and Miikkulainen]{BAIKAL}
Gonzalez, S. and Miikkulainen, R.
\newblock Improved training speed, accuracy, and data utilization through loss function optimization.
\newblock In \emph{2020 IEEE Congress on Evolutionary Computation (CEC)}, pp.\  1--8, 2020.
\newblock \doi{10.1109/CEC48606.2020.9185777}.

\bibitem[Gonzalez \& Miikkulainen(2021)Gonzalez and Miikkulainen]{LossTaylor}
Gonzalez, S. and Miikkulainen, R.
\newblock Optimizing loss functions through multi-variate {Taylor} polynomial parameterization.
\newblock In \emph{Proceedings of the Genetic and Evolutionary Computation Conference}, GECCO '21, pp.\  305–313, New York, NY, USA, 2021. Association for Computing Machinery.
\newblock ISBN 9781450383509.
\newblock \doi{10.1145/3449639.3459277}.
\newblock URL \url{https://doi.org/10.1145/3449639.3459277}.

\bibitem[Gu et~al.(2022)Gu, Li, Fu, Wong, Chen, and Zhu]{SoftmaxPersonLoss}
Gu, H., Li, J., Fu, G., Wong, C., Chen, X., and Zhu, J.
\newblock {AutoLoss-GMS}: Searching generalized margin-based softmax loss function for person re-identification.
\newblock In \emph{Proceedings of the IEEE/CVF Conference on Computer Vision and Pattern Recognition (CVPR)}, pp.\  4744--4753, June 2022.

\bibitem[He et~al.(2016)He, Zhang, Ren, and Sun]{ResNetV2}
He, K., Zhang, X., Ren, S., and Sun, J.
\newblock Identity mappings in deep residual networks.
\newblock In \emph{European Conference on Computer Vision}, pp.\  630--645. Springer, 2016.

\bibitem[He et~al.(2021)He, Zhao, and Chu]{AUTOML}
He, X., Zhao, K., and Chu, X.
\newblock {AutoML}: A survey of the state-of-the-art.
\newblock \emph{Knowledge-Based Systems}, 212:\penalty0 106622, 2021.

\bibitem[Hu et~al.(2018)Hu, Shen, and Sun]{SERESNET}
Hu, J., Shen, L., and Sun, G.
\newblock Squeeze-and-excitation networks.
\newblock In \emph{Proceedings of the IEEE Conference on Computer Vision and Pattern Recognition}, pp.\  7132--7141, 2018.

\bibitem[Kendall(1938)]{Kendall}
Kendall, M.~G.
\newblock A new measure of rank correlation.
\newblock \emph{Biometrika}, 30\penalty0 (1/2):\penalty0 81--93, 1938.
\newblock ISSN 00063444.
\newblock URL \url{http://www.jstor.org/stable/2332226}.

\bibitem[Kingma \& Ba(2014)Kingma and Ba]{Adam}
Kingma, D.~P. and Ba, J.
\newblock Adam: A method for stochastic optimization.
\newblock \emph{arXiv preprint arXiv:1412.6980}, 2014.

\bibitem[Krizhevsky \& Hinton(2009)Krizhevsky and Hinton]{CIFAR}
Krizhevsky, A. and Hinton, G.
\newblock Learning multiple layers of features from tiny images.
\newblock Technical report, University of Toronto, Toronto, Ontario, 2009.

\bibitem[Leng et~al.(2022)Leng, Tan, Liu, Cubuk, Shi, Cheng, and Anguelov]{TaylorNPoly}
Leng, Z., Tan, M., Liu, C., Cubuk, E.~D., Shi, X., Cheng, S., and Anguelov, D.
\newblock Polyloss: A polynomial expansion perspective of classification loss functions.
\newblock \emph{arXiv preprint}, 2022.
\newblock \doi{10.48550/ARXIV.2204.12511}.
\newblock URL \url{https://arxiv.org/abs/2204.12511}.

\bibitem[Li et~al.(2022)Li, Fu, Dai, Li, Huang, and Zhu]{AutoLoss}
Li, H., Fu, T., Dai, J., Li, H., Huang, G., and Zhu, X.
\newblock Autoloss-zero: Searching loss functions from scratch for generic tasks.
\newblock In \emph{2022 IEEE/CVF Conference on Computer Vision and Pattern Recognition (CVPR)}, pp.\  999--1008, 2022.
\newblock \doi{10.1109/CVPR52688.2022.00108}.

\bibitem[Li(2017)]{CodeResNet}
Li, W.
\newblock cifar-10-cnn: Play deep learning with {CIFAR} datasets.
\newblock \url{https://github.com/BIGBALLON/cifar-10-cnn}, 2017.

\bibitem[Li(2022)]{CodeEffNet}
Li, W.
\newblock Keras {EfficientNetV2}.
\newblock \url{https://github.com/leondgarse/keras_efficientnet_v2}, 2022.

\bibitem[Liu et~al.(2020)Liu, Brock, Simonyan, and Le]{BNGA}
Liu, H., Brock, A., Simonyan, K., and Le, Q.
\newblock Evolving normalization-activation layers.
\newblock \emph{Advances in Neural Information Processing Systems}, 33:\penalty0 13539--13550, 2020.

\bibitem[Liu et~al.(2021)Liu, Zhang, Wang, Xu, Liang, Jiang, and Li]{LossObjectDetection}
Liu, P., Zhang, G., Wang, B., Xu, H., Liang, X., Jiang, Y., and Li, Z.
\newblock Loss function discovery for object detection via convergence-simulation driven search.
\newblock \emph{arXiv preprint arXiv:2102.04700}, 2021.

\bibitem[Luke \& Panait(2006)Luke and Panait]{BLOAT}
Luke, S. and Panait, L.
\newblock A comparison of bloat control methods for genetic programming.
\newblock \emph{Evolutionary Computation}, 14\penalty0 (3):\penalty0 309--344, 2006.

\bibitem[Neshatian \& Zhang(2009)Neshatian and Zhang]{REDUNDANCY}
Neshatian, K. and Zhang, M.
\newblock Unsupervised elimination of redundant features using genetic programming.
\newblock In Nicholson, A. and Li, X. (eds.), \emph{AI 2009: Advances in Artificial Intelligence}, pp.\  432--442, Berlin, Heidelberg, 2009. Springer Berlin Heidelberg.
\newblock ISBN 978-3-642-10439-8.

\bibitem[Raymond et~al.(2022)Raymond, Chen, Xue, and Zhang]{LearnSymbLossMeta}
Raymond, C., Chen, Q., Xue, B., and Zhang, M.
\newblock Learning symbolic model-agnostic loss functions via meta-learning.
\newblock \emph{arXiv preprint arXiv:2209.08907}, 2022.

\bibitem[Real et~al.(2019)Real, Aggarwal, Huang, and Le]{AmoebaNet}
Real, E., Aggarwal, A., Huang, Y., and Le, Q.~V.
\newblock Regularized evolution for image classifier architecture search.
\newblock In \emph{Proceedings of the AAAI Conference on Artificial Intelligence}, volume~33, pp.\  4780--4789, 2019.

\bibitem[Stanley \& Miikkulainen(2002)Stanley and Miikkulainen]{NEAT}
Stanley, K.~O. and Miikkulainen, R.
\newblock Evolving neural networks through augmenting topologies.
\newblock \emph{Evolutionary Computation}, 10\penalty0 (2):\penalty0 99--127, 2002.

\bibitem[Tan \& Le(2021)Tan and Le]{EffNetV2}
Tan, M. and Le, Q.
\newblock {Efficientnetv2}: Smaller models and faster training.
\newblock In \emph{International Conference on Machine Learning}, pp.\  10096--10106. PMLR, 2021.

\bibitem[Yun et~al.(2019)Yun, Han, Oh, Chun, Choe, and Yoo]{CutMix}
Yun, S., Han, D., Oh, S.~J., Chun, S., Choe, J., and Yoo, Y.
\newblock Cutmix: Regularization strategy to train strong classifiers with localizable features.
\newblock In \emph{Proceedings of the IEEE/CVF International Conference on Computer Vision}, pp.\  6023--6032, 2019.

\bibitem[Zagoruyko \& Komodakis(2016)Zagoruyko and Komodakis]{WideResNet}
Zagoruyko, S. and Komodakis, N.
\newblock Wide residual networks.
\newblock \emph{arXiv preprint arXiv:1605.07146}, 2016.

\bibitem[Zoph \& Le(2016)Zoph and Le]{OGNASNet}
Zoph, B. and Le, Q.~V.
\newblock Neural architecture search with reinforcement learning.
\newblock \emph{arXiv preprint}, abs/1611.01578, 2016.
\newblock \doi{10.48550/ARXIV.1611.01578}.
\newblock URL \url{https://arxiv.org/abs/1611.01578}.

\bibitem[Zoph et~al.(2018)Zoph, Vasudevan, Shlens, and Le]{NASNet}
Zoph, B., Vasudevan, V., Shlens, J., and Le, Q.~V.
\newblock Learning transferable architectures for scalable image recognition.
\newblock In \emph{Proceedings of the IEEE Conference on Computer Vision and Pattern Recognition (CVPR)}, June 2018.

\end{thebibliography}
%%
%% If your work has an appendix, this is the place to put it.
\appendix

\section{Algorithms}
\label{app:algos}
Pseudocode for the integrity check and the genetic algorithm are given in Algorithm~\ref{alg:integrity} and Algorithm~\ref{alg:ga}, respectively.

\begin{algorithm}[htb]
   \caption{Integrity Check}
   \label{alg:integrity}
\begin{algorithmic}[1]
   \STATE {\bfseries Input:} Candidate Loss $loss$, Population $population$
    \IF{ContainsCycle($loss$) == True}
    \STATE Return False
   \ENDIF
   \IF{ContainsYandYHat($loss$) == False}
    \STATE Return False
   \ENDIF
   \IF{ContainsOneGlobalMinOrMax($loss$) == False}
    \STATE Return False
   \ENDIF
    \STATE CheckSign($loss$)
   \FOR{$individual$ {\bfseries in} $population$}
   \IF{Norm($individual-loss$)$<0.01$}
   \STATE Return False
   \ENDIF
   \ENDFOR
    \STATE Return True
\end{algorithmic}
\end{algorithm}

\begin{algorithm}[htb]
   \caption{Genetic Algorithm}
   \label{alg:ga}
\begin{algorithmic}[1]
    \STATE $initialPopulation$ = LossInitialization(1000)
    \STATE $validationAcc$ = ConvNet($initialPopulation$)
    \STATE $population$ = Best100($initialPopulation$, $validationAcc$)
    \REPEAT
    \STATE $winner$ = Tournament($population$, $validationAcc$, 20)
    \STATE $maxRedo$ = 7
    \WHILE{$maxRedo$ $>$ 0 }
    \STATE $maxMutation$ = 2500
    \WHILE{$maxMutation$ $>$ 0 }
    \STATE $child$ = Mutation($winner$)
    \IF{$child$.CheckIntegrity()}
    \STATE break inner while loop
    \ELSE
    \STATE $maxMutation$ --= 1
    \ENDIF
    \ENDWHILE
    \STATE $valAcc$ = TrainEffNetV2($child$)
    \IF{$valAcc$ == None}
    \STATE $maxRedo$ --= 1
    \ELSE
    \STATE break while loop
    \ENDIF
    \ENDWHILE
    \IF{$maxRedo$ == 0}
    \STATE Pass 
    \ELSE
    \STATE $population$ = ReplaceOldest($population$, $child$)
    \STATE $validationAcc$ = Update($validationAcc$, $valAcc$)
    \ENDIF
    \UNTIL{Time Expired}
\end{algorithmic}
\end{algorithm}

\section{Further Mutation Details}
\label{app:mutation}
Mutation is performed by selecting either the root node or one of the hidden state nodes uniformly randomly. Once a node is chosen, either the operation or connection is changed. With 70\% probability, the operation is changed either from unary to unary, or binary to binary, depending on the original operation. With 15\% probability, the operation was swapped from unary to binary, or binary to unary, depending upon the original operation. If the original operation was unary, the extra connection obtained from swapping to binary was randomly selected from the set of input and hidden state nodes. If the original operation was binary, then either the left or right connection was eliminated with 50\% probability. If the operation was not changed or swapped, the remaining 15\% probability went to changing the connection of the node. If the operation was unary, then the connection was changed uniformly random from the set of other possible connections. If the operation was binary, then with 20\% probability the argument connections were swapped; else, either were changed uniformly randomly. 

The child loss was then trained using the surrogate function. Despite the integrity check performed when creating the loss functions, losses can still not learn during training. To prevent wasting computation, within the surrogate function an early stopping mechanism was incorporated to stop the learning process for degenerate losses. If this occurred, another child loss was obtained through mutation. If seven mutated child loss functions in a row failed integrity checks, the mutation procedure was considered to be failed, and another iteration of the algorithm was performed. The entire genetic algorithm used in this work is detailed in Algorithm~\ref{alg:ga}.

\section{Further Implementation Details}
\label{app:implementation}

All architectures, training strategies, and loss functions were implemented in Tensorflow \citep{Tensorflow}. The code base for the EffNetV2Small model and RandAug regularization strategy were derived from the repository \citep{CodeEffNet}. The code base for WideResNet and SEResNeXt were derived from the repository \citep{CodeResNet}. All of our code is freely accessible through our repository \citep{CodeNeural}. For computational resources, we had access to a single NVIDIA A100 GPU during evolution. 

\section{Training Strategies}
\label{app:train_strat}

Table~\ref{tab:randaug} shows the adapted progressive RandAug regularization training strategy for CIFAR-10 and CIFAR-100 for EffNetV2Small. The surrogate function was trained for only 8,000 steps of phase~1. The proposed loss function elimination protocol trained on each phase incrementally, where each phase contained 16,000 steps. EffNetV2Small was always trained using the Adam optimizer \citep{Adam} with a maximum learning rate of 0.001 and a batch size of 64. The first 6,400 steps were used to warm up the learning rate linearly from 0 to the maximum, and then the learning rate was decayed using cosine annealing back down to 0 during the rest of the training cycle. This training strategy was only used for the surrogate function and loss function elimination protocol.

\begin{table}[bht]
\caption{Adapted progressive RandAug regularization training strategy for CIFAR-10 and CIFAR-100.}
\label{tab:randaug}
\vskip 0.1in
\begin{center}
\begin{small}
\begin{sc}
\begin{tabular}{ p{1.6cm} || p{1.12cm}p{1.12cm}p{1.12cm}p{1.12cm}}
\toprule
 Phase &  1 &  2 &  3 &  4 \\
\midrule
Image Size & 128 & 160 & 192 & 224 \\ 
Dropout & 0.10 & 0.20 & 0.30 & 0.40 \\ 
RandAug & 5 & 8 & 12 & 15 \\ 
\bottomrule
\end{tabular}
\end{sc}
\end{small}
\end{center}
\vskip -0.1in
\end{table}

When comparing the final loss functions after evolution and the elimination protocol in Section~\ref{sec:results}, the training strategy was adjusted for EffNetV2Small to allow for a longer training session. Each phase was increased to 24,000 steps, allowing for 96,000 total steps over the entire course of training. The same optimizer and learning rate strategy were used as before, except the learning rate was held at its maximum of 0.001 for 12,400 steps before being decayed. 

For WideResNet and SEResNeXt, SGD with Nesterov momentum of 0.9 was used with a learning rate scheduler that warmed up to 0.1 after 1,000 steps, and then was cosine decayed back down to 0. This same training strategy was used by \citet{WideResNet} for the WideResNet architecture, except they decayed their learning rate with a piece-wise function rather than cosine. With a batch size of 128, a total of 96,186 steps were performed. For image augmentation, the images were resized to 40$\times$40, randomly cropped back to 32$\times$32, cutout was applied with an 8$\times$8 mask, and then the images were randomly flipped horizontally. 

For DavidNet, two different image augmentation techniques were applied. For both, SGD with Nesterov momentum of 0.9 was used with a learning rate scheduler that warmed up to 0.4 after 1,176 steps, and then was linearly decayed. For the first configuration using the random crop, cutout, and horizontal flipping, a weight decay value of 0.512 was also applied to the gradients. This training strategy was used by the original DavidNet proposal for the DAWNBench competition \citep{DawnBench}, an end-to-end deep learning benchmark for training time, cost, and accuracy for CIFAR-10 and ImageNet. With a batch size of 512, 98 steps were performed per epoch, for 30 epochs, totalling to a small budget of 2,940 steps. As a result, the loss functions are evaluated for whether or not they can increase generalization within a very computationally stringent budget. For the second image augmentation technique, CutMix, the exact same parameters and setup were used for DavidNet, except the weight decay was reduced to 0.0256, as CutMix is a much harsher augmentation technique than simple cutout.

When training EffNetV2Small on Stanford-CARS196, Oxford-Flowers102, and Caltech101 from scratch, a few minor changes were in-placed. The same optimizer, learning rate, schedule, and total steps were used as before; however, the progressive RandAug regularization strategy was reverted back to its original form \citep{EffNetV2} for the much larger image resolution datasets. These changes are reflected in Table~\ref{tab:randaug2}. Due to the increased image resolution, the batch sizes for EffNetV2Small were decreased to 32.

When fine-tuning, the initial weights were set to the official ImageNet2012 weights, and the loss functions were trained for only 20 epochs, with 500 steps per epoch, totalling to only 10,000 steps. The images were set to 224x224, dropout to 0.5, RandAug magnitude to 15, batch size to 64, ptimizer to Adam, and learning rate schedule that warmed up for one epoch to 0.001 before being decayed back down to zero after being held for three epochs. 

\begin{table}[htb]
\caption{Original RandAug progressive regularization training strategy for EffNetV2Small for large resolution image datasets.}
\label{tab:randaug2}
\vskip 0.1in
\begin{center}
\begin{small}
\begin{sc}
\begin{tabular}{ p{1.6cm} || p{1.12cm}p{1.12cm}p{1.12cm}p{1.12cm}}
\toprule
Phase & 1 & 2 & 3 & 4 \\
\midrule
Image Size & 128 & 186 & 244 & 300 \\ 
Dropout & 0.10 & 0.17 & 0.24 & 0.30 \\ 
RandAug & 5 & 8 & 12 & 15 \\ 
\bottomrule
\end{tabular}
\end{sc}
\end{small}
\end{center}
\vskip -0.1in
\end{table}

\end{document}